\documentclass[letterpaper, 10pt, conference]{ieeeconf}  

\IEEEoverridecommandlockouts                              

\usepackage{graphics} 
\usepackage{amsmath,amssymb,amsfonts}
\usepackage{graphicx}
\usepackage{hyperref}
\usepackage{booktabs}
\usepackage{tabularx}
\usepackage{subcaption}
\usepackage{siunitx}
\usepackage{makecell}

\title{\LARGE \bf
Unveiling Objects with SOLA: An Annotation-Free Image Search on the Object Level for Automotive Data Sets
}

\author{Philipp Rigoll$^{1}$, Jacob Langner$^{1}$ and Eric Sax$^{2}$
\thanks{$^{1}$Philipp Rigoll and Jacob Langner are with FZI Research Center for Information Technology, 76131 Karlsruhe, Germany
        {\tt\small philipp.rigoll@fzi.de, langner@fzi.de}}
\thanks{$^{2}$Eric Sax is with Karlsruhe Institute of Technology, 76131 Karlsruhe, Germany
        {\tt\small eric.sax@kit.edu}}%
}

\begin{document}

\maketitle
\thispagestyle{empty}
\pagestyle{empty}

\begin{abstract}
Huge image data sets are the foundation for the development of the perception of automated driving systems.
A large number of images is necessary to train robust neural networks that can cope with diverse situations.
A sufficiently large data set contains challenging situations and objects.
For testing the resulting functions, it is necessary that these situations and objects can be found and extracted from the data set.
While it is relatively easy to record a large amount of unlabeled data, it is far more difficult to find demanding situations and objects.
However, during the development of perception systems, it must be possible to access challenging data without having to perform lengthy and time-consuming annotations.
A developer must therefore be able to search dynamically for specific situations and objects in a data set.
Thus, we designed a method which is based on state-of-the-art neural networks to search for objects with certain properties within an image.
For the ease of use, the query of this search is described using natural language.
To determine the time savings and performance gains, we evaluated our method qualitatively and quantitatively on automotive data sets.
\end{abstract}

\section{INTRODUCTION}
A robust perception is key for reliable automated driving systems (ADS).
These perception algorithms must work flawlessly under extremely diverse situations.
To ensure a proper behavior, it must be ensured that the test data sets for these algorithms contain the relevant situations.
We focus on camera images, because they are an important input to the perception algorithms.
Thus, it is necessary to extract specific images from a data set during the development of an ADS in order to develop and test with them.
In contrast to structured data, it is not trivial to search for specific images.
One option is to manually search and annotate the entire image database.
However, not all problematic situations are known beforehand.
Some problems come to light during development and data sets would have to be dynamically annotated.
Furthermore, the cost of annotation increases with the size of the data set~\cite{chen_beat_2014}.
Therefore, we utilize pretrained neural networks and present a method that does not require annotated data.

\begin{figure}%
	\centering
	\includegraphics[width=.9\columnwidth]{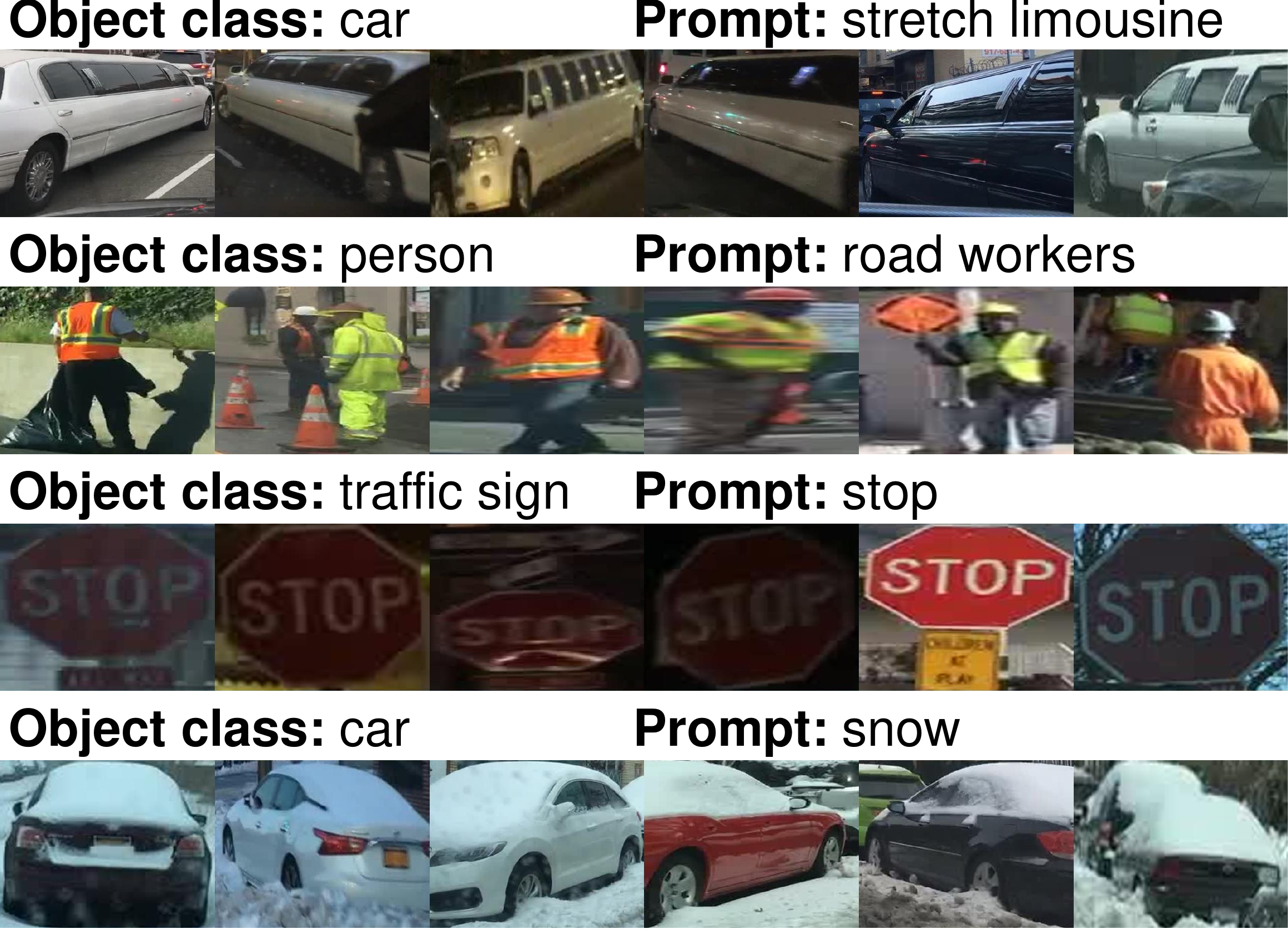}%
	\caption{Examples of specific objects found with our method (SOLA) in the BDD100k data set~\cite{yu_bdd100k_2020}.}%
	\label{fig:pull_image}%
\end{figure}

In our previous work~\cite{rigoll_focus_2023} we already investigated a method to search for images based on a textual description of the full image. 
However, this method focuses on searching a full image and therefore the capability of searching for individual objects within an image is limited.
Now we want to enable the targeted search for objects with certain properties within the image, even if the objects are not dominant in the image and may only make up a part of the full image.
For example, our goal is to search for police men, vehicle brands, and specific traffic sign types (see Fig.~\ref{fig:pull_image}).

Hence, we built upon state-of-the-art neural networks in order to allow a detailed search in images on an object level.
This resulted in our annotation-free image \textbf{s}earch method on the \textbf{o}bject \textbf{l}evel for \textbf{a}utomotive data sets (SOLA).
The focus lies on a powerful and easy to use search option.
No programming knowledge is required to use SOLA, and the search is performed using natural language.
To test the performance of SOLA, we evaluate it extensively on automotive data sets.

The structure of our paper is as follows: in Section~\ref{sec:related_work} we analyze the current work in the field of object retrieval and give a small insight into the functioning of a state-of-the-art neural network, which is an important module in SOLA.
SOLA was developed starting from requirements.
These requirements and the exact structure and procedure of SOLA are explained in section~\ref{sec:method}.
Several detailed qualitative and quantitative evaluations are presented in Section~\ref{sec:evaluation}.
In Section~\ref{sec:conclusion}, we summarize the main points of our paper and preview future research topics.

\section{RELATED WORK}
\label{sec:related_work}

A decisive capability of Automotive Systems Engineering~\cite{petersen_towards_2022} is mastering the data management~\cite{luckow_automotive_2015}.
The challenge here is the size of the data sets~\cite{bogdoll_ad-datasets_2022,yin_when_2017}.
During the development of the perception of ADS, it is necessary to examine its behavior in all relevant situations~\cite{li_coda_2022}.
One important point are objects that rarely occur and for which the performance of the perception is unknown~\cite{oksuz_imbalance_2021,bogdoll_perception_2023}.

The procedure of identifying relevant images within a data set based on a given context is called image retrieval.
One approach is to label the images manually with tags, automatically complete the tagging of the whole data set and use these tags to retrieve images~\cite{lei_wu_tag_2013}.
Another option to find specific images without manual labeling is by looking at the context data like the recording location and the timestamp~\cite{rigoll_scalable_2022}.
Thus, geographic map data can be utilized in order to search for static objects.
And it is possible to search for environmental phenomena such as weather conditions and sun positions~\cite{qi_anomaly_2018}.
For a more dynamic search, it is possible to sketch the outlines of a desired image~\cite{lin_zero-shot_2023}.
At the individual image level, it is possible to search for specific objects using natural language~\cite{hu_segmentation_2016, ding_vlt_2023, luddecke_image_2022, liu_gres_2023, liu_grounding_2023}.
However, these neural networks would have to process each image anew during every search.
Kirillov et al.~\cite{kirillov_segment_2023} partially prevent this problem by separating their time-consuming calculation of an image embedding so that only a subnetwork has to be executed during search.
In contrast, we investigate an approach where no inference is necessary for the image dataset during the search.
Goenka et al.~\cite{goenka_fashionvlp_2022} search with a reference image and natural language feedback.
Thereby, they focus on the full images of a fashion data set.

One of the neural networks we use to search an entire data set at object level is CLIP.
CLIP (Contrastive Language Image Pre-training) is a neural network which was developed by Radford et al.~\cite{radford_learning_2021-1}.
Contrastive learning is a technique used to learn feature embeddings where similar data points get mapped to similar representations while ensuring that dissimilar data points are mapped to representations that are far apart from each other.
CLIP is trained on 400~million image-text pairs, which were obtained by gathering images and their alternative text descriptions on the public internet.
The trained neural network consists of two encoders, which are able to transfer an image $\mathcal{E}_\text{image}$ or text $\mathcal{E}_\text{text}$ into a vector representation.
The text and the image vector representations are part of a shared latent space.
Similar images and similar texts are transformed to similar vector representations.

\section{METHOD}
\label{sec:method}

From the planned application of SOLA during the development of automated driving systems, we identified the following requirements:
\begin{itemize}
    \item no annotation expense,
    \item search for properties on an object level,
    \item easy to formulate queries,
    \item sensible runtime during application.
\end{itemize}

\begin{figure*}%
	\centering
	\includegraphics[width=.9\textwidth]{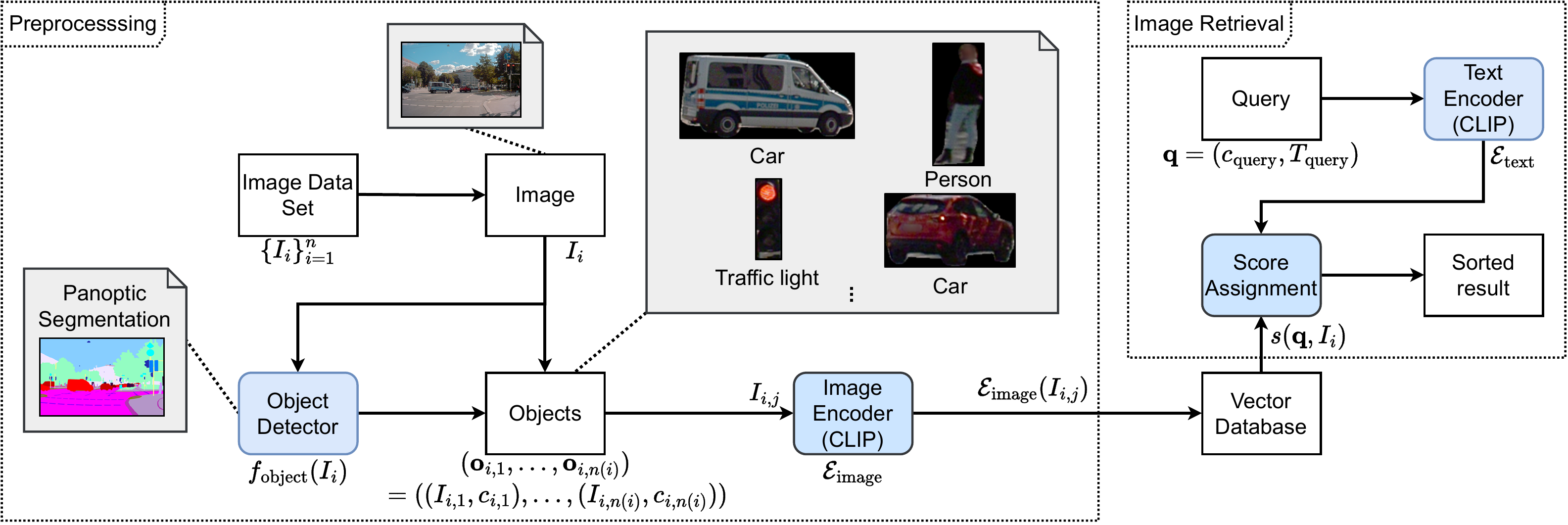}%
	\caption{Overview of the functionality of SOLA (example image from the A2D2 data set~\cite{geyer_a2d2_2020}).}%
	\label{fig:method}%
\end{figure*}

Therefore, we propose SOLA that aims to identify the most relevant images for a given query within an image data set $\{ I_i\}_{i=1}^n$.
The query $\mathbf{q}=(c_\text{query}, T_\text{query})$ consists of an object class indicator $c_\text{query}$, which can be chosen from a predefined set, and an arbitrary natural text description $T_\text{query}$ that describes the object being searched for.
We assign a score to each image based on its alignment with $\mathbf{q}$ and sort the images accordingly.
SOLA (see Fig.~\ref{fig:method}) is divided into a preprocessing and the subsequent image retrieval step.

\subsection{Preprocessing}
At first, the isolated objects of each image are required.
Therefore, an object detector $f_\text{object}$ extracts all instances contained in an image $I_i \in I$:
\begin{equation}
    f_\text{object}(I_i) = (\mathbf{o}_{i,1}, \dots, \mathbf{o}_{i,n(i)}) \text{ ,}
\end{equation}
where each object $\mathbf{o}_{i,j} = (I_{i,j}, c_{i,j})$ comprises a cropped view $I_{i,j}$ of $I_i$ and a predicted class label $c_{i,j}$.
Note, that the total number of detected objects $n(i)$ depends on the content of the image.

We use a panoptic segmentation as an object detector~$f_\text{object}$.
A panoptic segmentation assigns each pixel to an object class $c \in C$ and groups them to objects.
$C$ is the set of all possible classes defined by the concrete implementation of the panoptic segmentation.
In addition to a semantic segmentation, the panoptic segmentation distinguishes between individual instances of an object.
To extract an object from the image, we apply the object mask created by the panoptic segmentation.
Thus, the background is blackened so that only the isolated object remains visible.
Afterwards, the bounding box that encloses the object is used to extract it from the original image.
One example of the preprocessed data is shown in Table~\ref{tab:databaseexample}.
Additionally, the bounding box information of the objects could be saved.
This would allow searchability by object position and object size.

\begin{table}
	\centering
	\caption{Example of the data produced by the proposed preprocessing step of SOLA.}%
	\label{tab:databaseexample}%
	\begin{tabular}{cll}
		\toprule
		\textbf{\#Image} & \textbf{Object Class} & \textbf{Vector Representation}\\
		\midrule
		1 & \texttt{traffic sign} &  $\mathcal{E}_\text{image}(I_{1,1})$\\
		1 & \texttt{car} & $\mathcal{E}_\text{image}(I_{1,2})$\\
		1 & \texttt{person} & $\mathcal{E}_\text{image}(I_{1,3})$\\
		1 & \texttt{car} & $\mathcal{E}_\text{image}(I_{1,4})$\\
        $\cdots$ & $\cdots$ & $\cdots$\\
        2 & \texttt{car} & $\mathcal{E}_\text{image}(I_{2,1})$\\
        2 & \texttt{person} & $\mathcal{E}_\text{image}(I_{2,2})$\\
        $\cdots$ & $\cdots$ & $\cdots$\\
		\bottomrule
	\end{tabular}
\end{table}

Next, an image encoder $\mathcal{E}_\text{image}$ maps each image patch with an object $I_{i,j}$ to a $d$-dimensional feature representation $\mathcal{E}_\text{image}(I_{i,j}) \in \mathbb{R}^d$.
We use the CLIP~\cite{radford_learning_2021} encoders to calculate the vector representations.
To calculate the vector representation for the object images $I_{i,j}$, the CLIP encoder $\mathcal{E}_\text{image}$ requires the object images to be quadratic.
Therefore, the object images must be squared.
To maintain the aspect ratio and every part of an object, zero padding is applied for squaring.

\subsection{Image Retrieval}
First, a text encoder $\mathcal{E}_\text{text}$ maps the text description $T_\text{query}$ to the same latent space as the images.
These representations are then used to compute a score $s$ which is based on the cosine similarity~\cite{national_institute_of_standards_and_technology_nist_cosine_2023}, which focuses on the angle between the vector representations:
\begin{equation}
    s(\textbf{q},\mathbf{o}_{i,j})= 
\begin{cases}
    \frac{\mathcal{E}_\text{image}(I_{i,j})^T \mathcal{E}_\text{text}(T_\text{query})}{\lVert \mathcal{E}_\text{image}(I_{i,j}) \rVert \lVert \mathcal{E}_\text{text}(T_\text{query}) \rVert } & \text{if }c_{i,j} = c_\text{query}, \\
    -\infty              & \text{otherwise.}
\end{cases}
\end{equation}
These scores represent the degree of alignment between individual objects and the query.

To assign a resulting score for the query, we compute
\begin{equation}
    s(\textbf{q},I_i) = \max \limits_{j=1,\dots,n(i)} s(\mathbf{q}, \mathbf{o}_{i,j}) \text{ .}
\end{equation}
The result of SOLA consists of images sorted according to this resulting score.

One could fall back to the original method~\cite{rigoll_focus_2023} where the full image is searched with CLIP. 
In this case, the vector representations of the full images $\mathcal{E}_\text{image}(I_i)$ are calculated and used for the image retrieval.

\subsection{Runtime}
One requirement is to ensure a reasonable runtime, therefore we recommend the following.

The preprocessing can be parallelized for the images and must only be performed once for every image.
If images are added to the data set, only the newly added images must be processed.
To speed up the following search, it is advisable to use a vector database~\cite{wang_milvus_2021,johnson_billion-scale_2017} which is optimized for the comparison of many vectors.

To speed up the sorting in the image retrieval step, it is advisable to filter all images with the score $-\infty$ because they do not include the desired object.

\section{EVALUATION}
\label{sec:evaluation}
To study the performance and the time savings by SOLA, we conducted several qualitative and quantitative experiments on automotive data sets.
For the implementation of SOLA, we have chosen Mask2Former~\cite{cheng_masked-attention_2022} trained on the cityscapes data set~\cite{cordts_cityscapes_2016} for the panoptic segmentation.
This method also returns a confidence score for each detected object.
In order not to lose any objects that might be difficult to detect, filtering by this score was omitted.

\begin{figure}%
	\centering
	\includegraphics[width=.95\columnwidth]{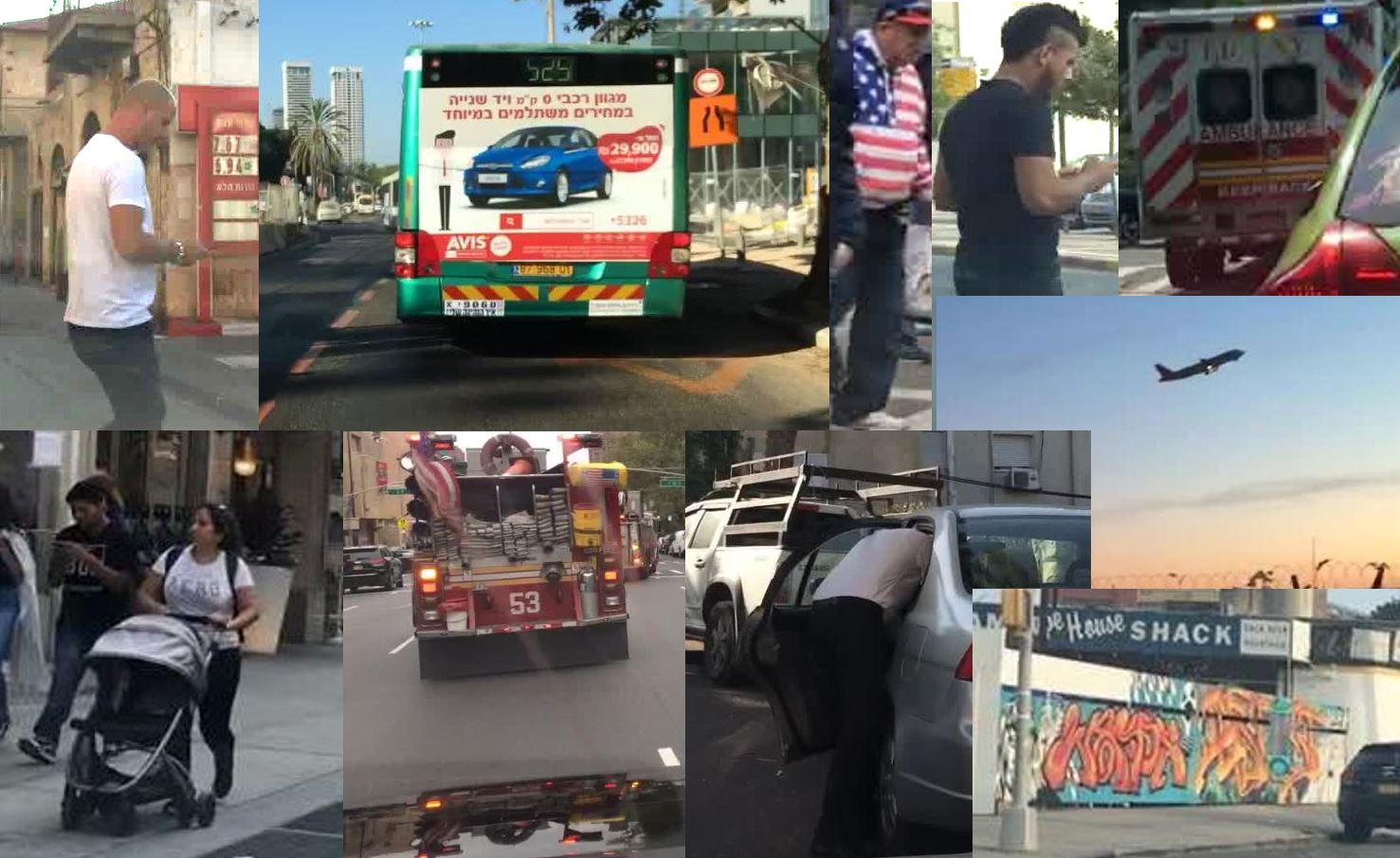}%
	\caption{Exemplary search for particularly rare objects (for better representation, partially cut out larger and brightened if necessary). Queries from left-to-right and top-to-bottom $(c_\text{query}, T_\text{query})$: (person, distracted), (bus, advertising with image of a car), (person, america), (person, distracted), (car, ambulance), (person, stroller), (truck, fire department), (person, bend over), (sky, airplane), (wall, graffiti).}%
	\label{fig:qualitative_example}%
\end{figure}

First, we started with a qualitative evaluation on the BDD100k data set~\cite{yu_bdd100k_2020}.
The BDD100k data set consists of $\SI{100000}{}$ images recorded during $\SI{50000}{}$ rides at diverse weather and road conditions.
For the qualitative evaluation, we thought of particular rare objects and tried to find these with appropriate queries within the first 10 results of SOLA.
Some exemplary results of this experiment are shown in Figure~\ref{fig:qualitative_example}.
One limitation that we have become aware of is that a targeted search is only possible for object classes that are also recognized by the panoptic segmentation.
For example, in our case it is not possible to search for animals because they are not recognized by Mask2Former.

\begin{figure*}%
	\centering
	\includegraphics[width=.5\textwidth]{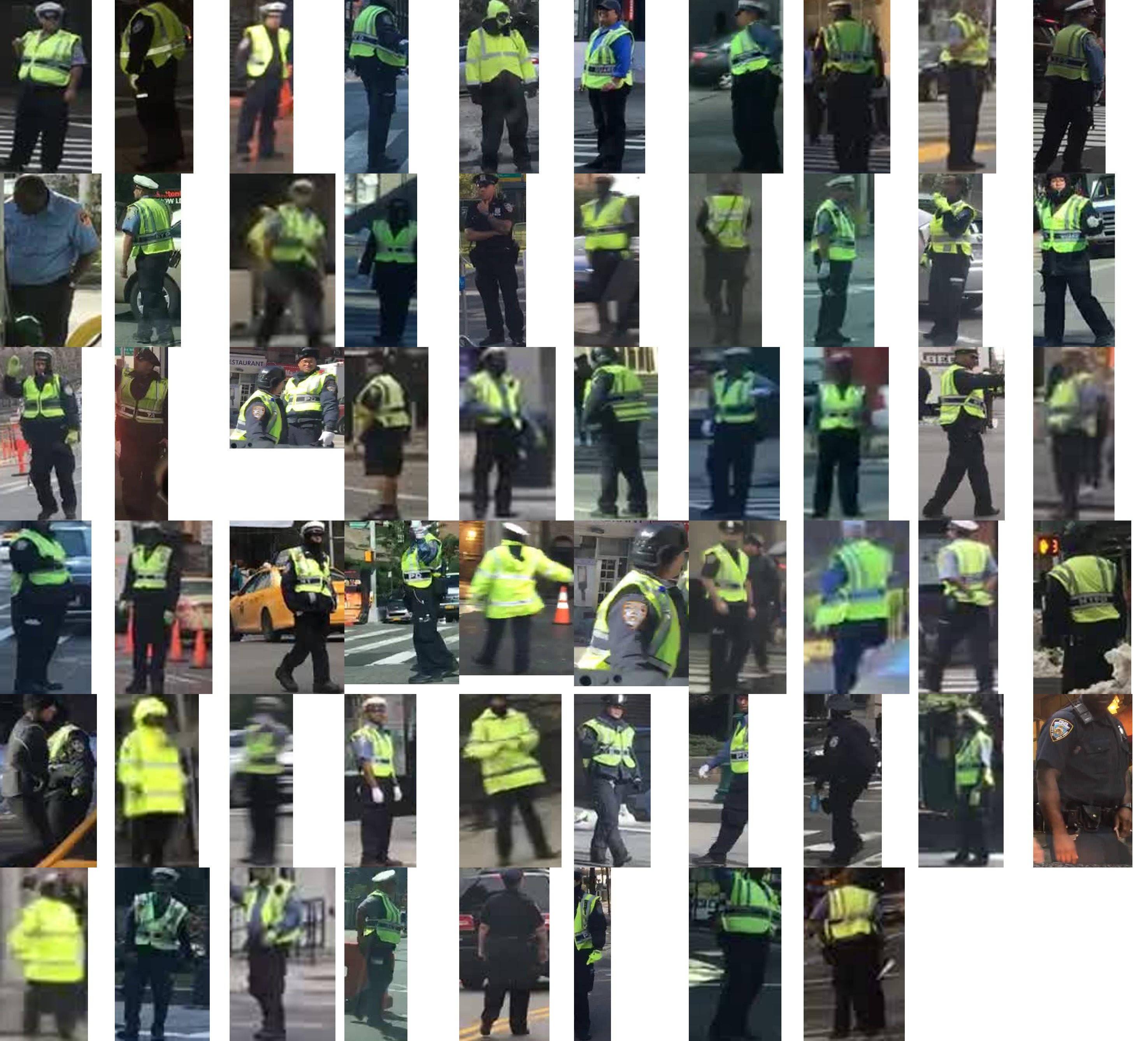}%
	\caption{Example image cutouts from true positive results found by SOLA with the query: object class ``person'' and prompt~``police man''.}%
	\label{fig:example_police_man}%
\end{figure*}

Next, we wanted to assess the performance of CLIP for the differentiation of individual object characteristics within an object class.
Therefore, we conducted experiments on the Stanford Cars data set~\cite{stanford_cars}, which comprises close-up views of approximately~200 car types.
The CLIP model demonstrated a classification accuracy exceeding~$\SI{90}{\percent}$, confirming its aptitude for processing automotive data.
Classification results for different prompt templates are reported in Table~\ref{tab:stanford_cars_classification}.
We concluded that we did not need to use a special prompt template for the further experiments.
We also examined BLIP~\cite{li_blip_2022} as a variation of CLIP, but were unable to achieve better results.

\begin{table}
	\centering
	\caption{Zero shot classification accuracy for different prompt templates evaluated on the Stanford Cars data set~\cite{stanford_cars} where the labels were the vehicle types.}%
	\label{tab:stanford_cars_classification}%
	\begin{tabular}{lcc}
		\toprule
		  \textbf{Prompt template} & \textbf{CLIP} & \textbf{BLIP}\\
		\midrule
        \texttt{\{label\}} & 89.1 & 53  \\
        \texttt{a \{label\}} & 88.7 & 52.3 \\
		\texttt{a photo of the nice \{label\}} & \textbf{91.4} & \textbf{55.3}\\
        \texttt{a picture of a \{label\}} & 91.3 & 54.5 \\
        \texttt{a low resolution photo of a \{label\}} & 90.4 & 52.5 \\
        \texttt{a bright photo of the \{label\}} & 87.4 & 53.2 \\
        \texttt{a pixelated photo of a \{label\}} & 85.8 & 51.1 \\
        \texttt{a jpeg corrupted photo of \{label\}} & 83.3 & 49.1 \\
		\bottomrule
	\end{tabular}
\end{table}

\begin{figure*}
    \centering
    \begin{subfigure}[t]{0.32\textwidth}
        \centering
        \includegraphics[width=\linewidth]{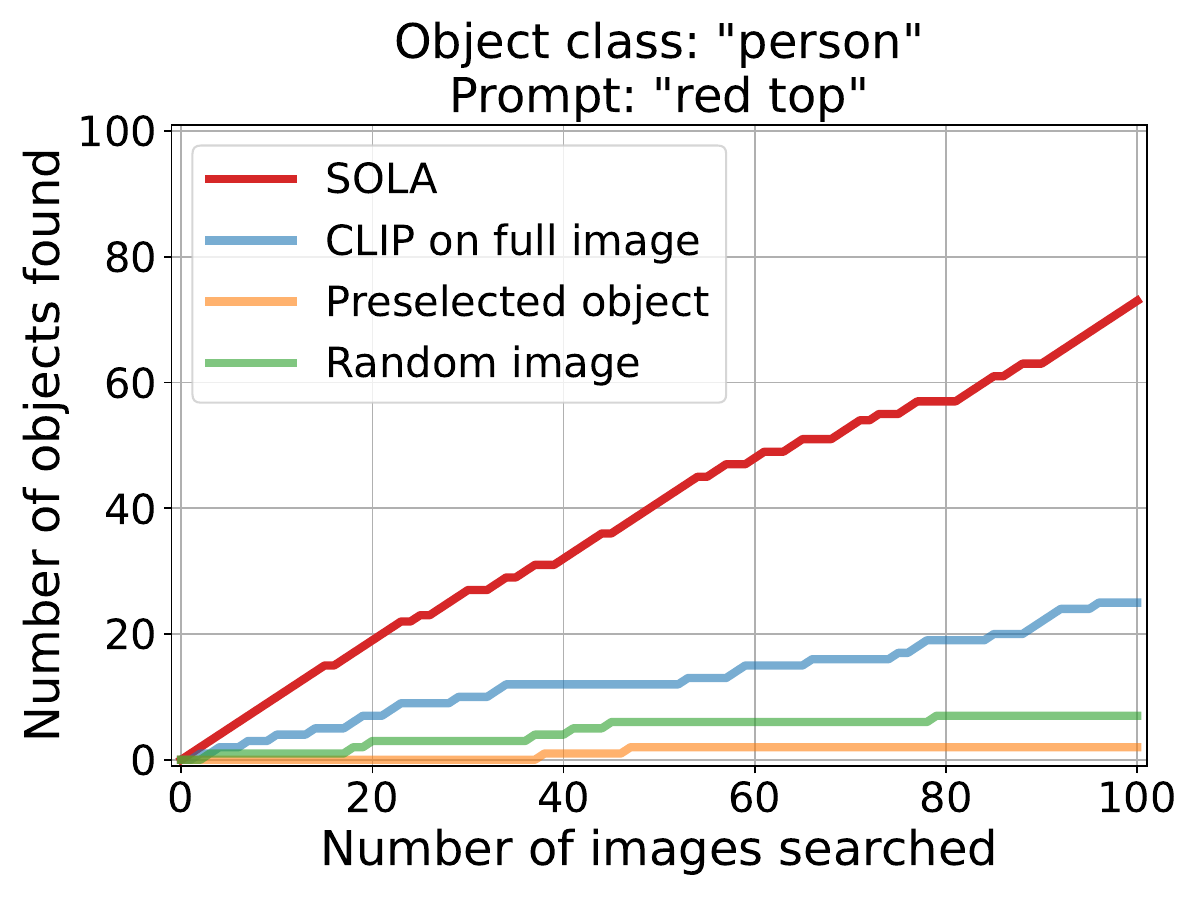}
        \label{fig:image_search_experiment:a}
    \end{subfigure}
    \hfill
    \begin{subfigure}[t]{0.32\textwidth}
        \centering
        \includegraphics[width=\linewidth]{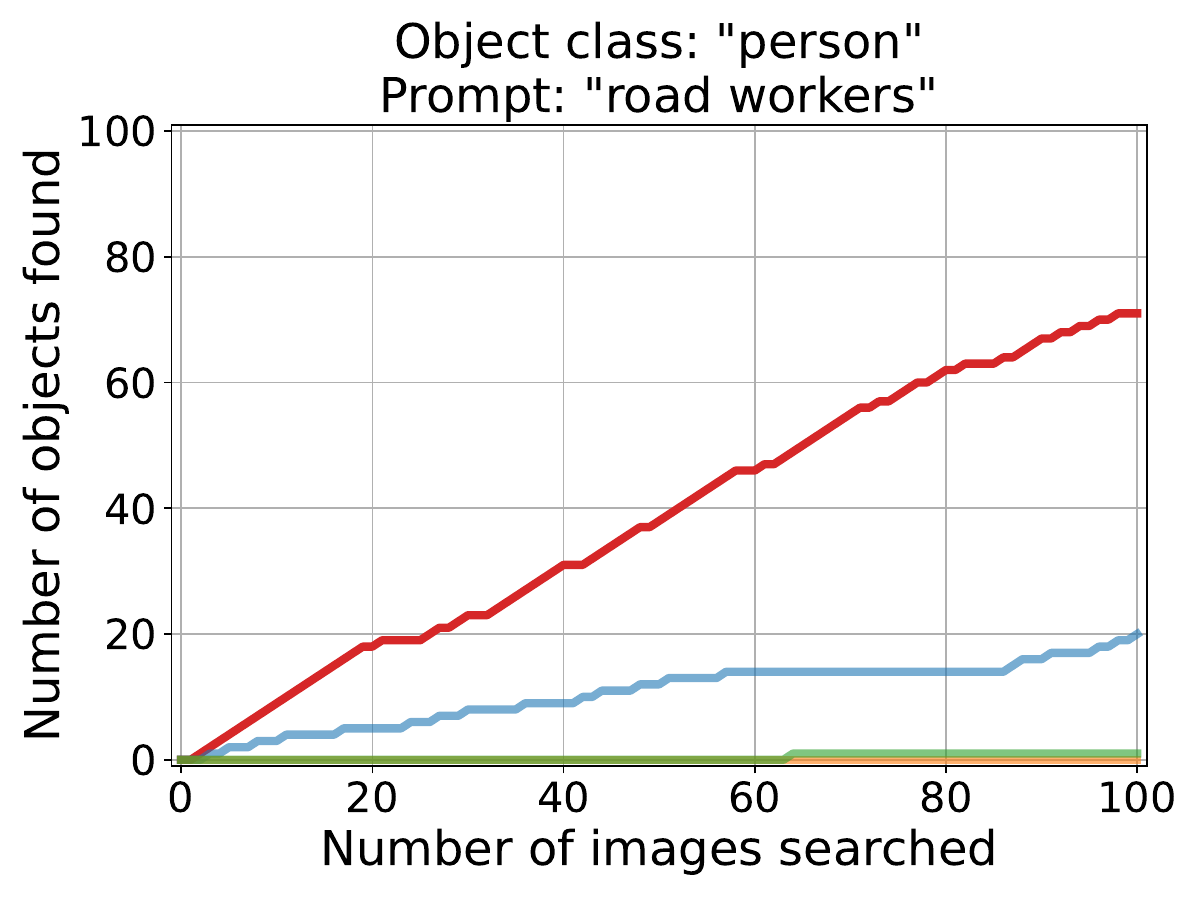}
        \label{fig:image_search_experiment:b}
    \end{subfigure}
    \hfill
    \begin{subfigure}[t]{0.32\textwidth}
        \centering
        \includegraphics[width=\linewidth]{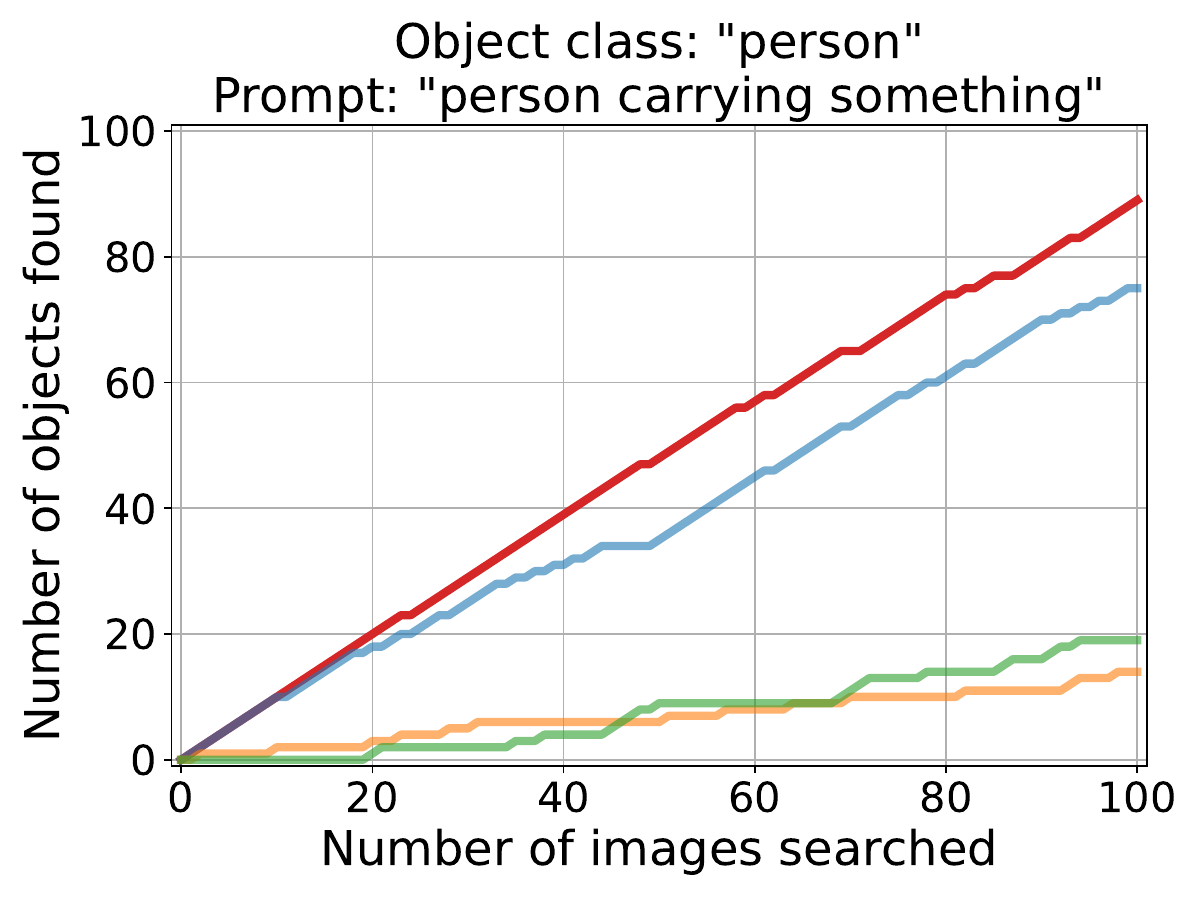}
        \label{fig:image_search_experiment:c}
    \end{subfigure}
    \hfill
    \begin{subfigure}[t]{0.32\textwidth}
        \centering
        \includegraphics[width=\linewidth]{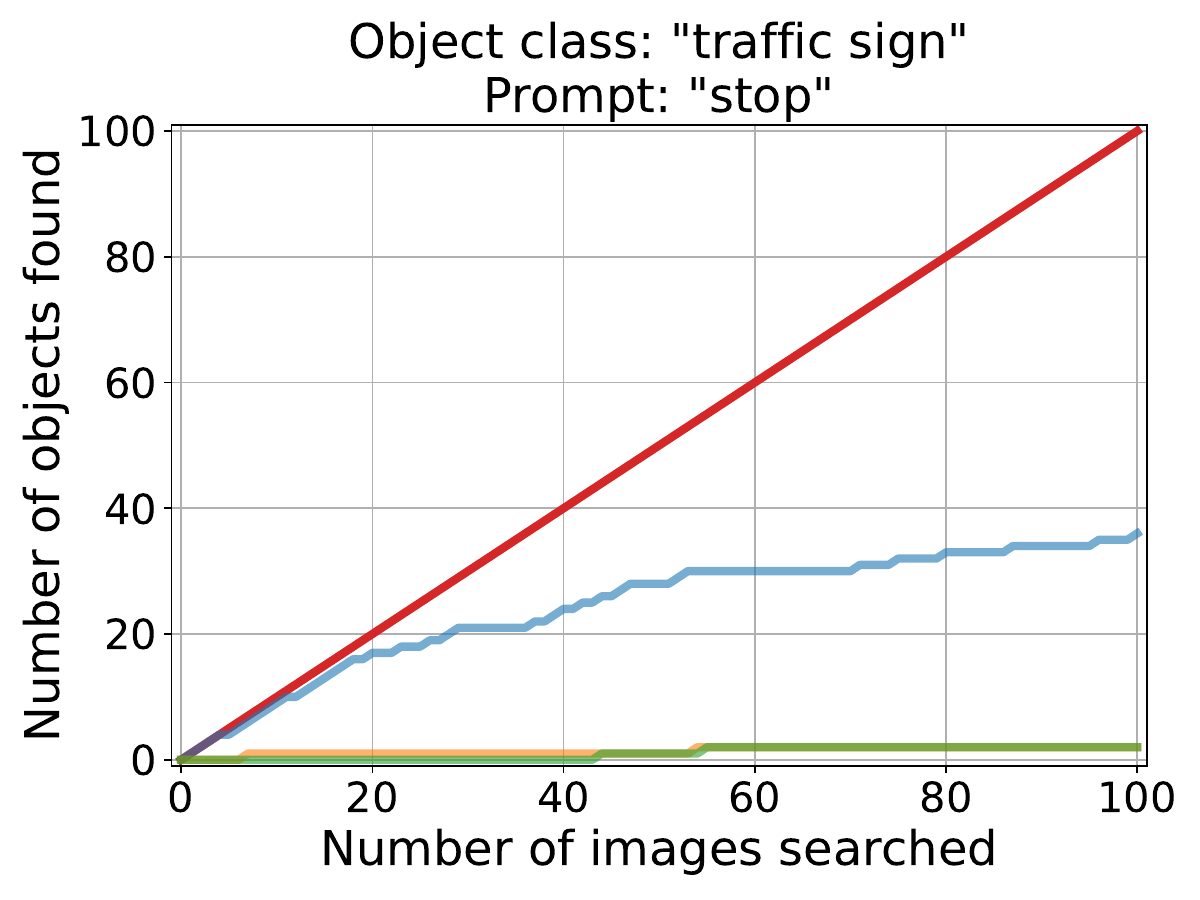}
        \label{fig:image_search_experiment:d}
    \end{subfigure}
    \hfill
    \begin{subfigure}[t]{0.32\textwidth}
        \centering
        \includegraphics[width=\linewidth]{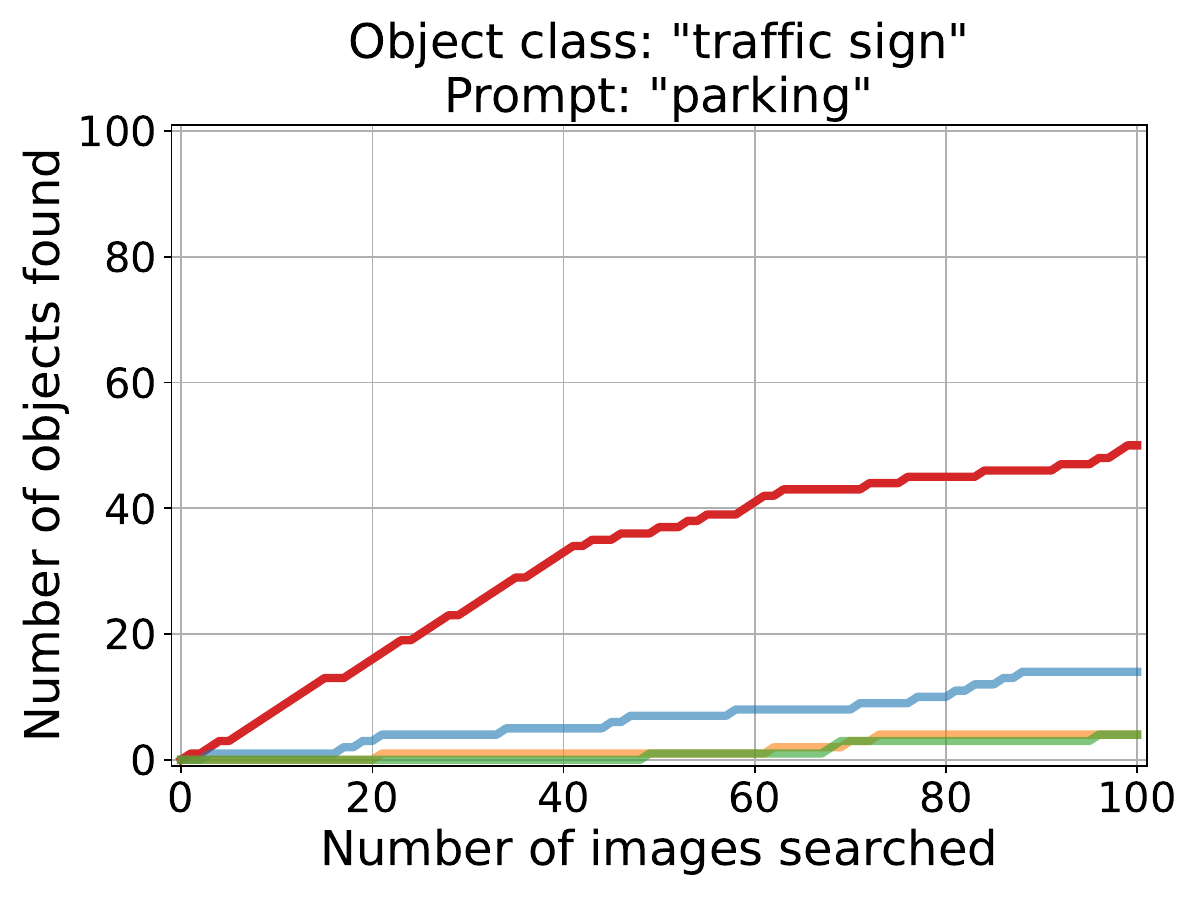}
        \label{fig:image_search_experiment:e}
    \end{subfigure}
    \hfill
    \begin{subfigure}[t]{0.32\textwidth}
        \centering
        \includegraphics[width=\linewidth]{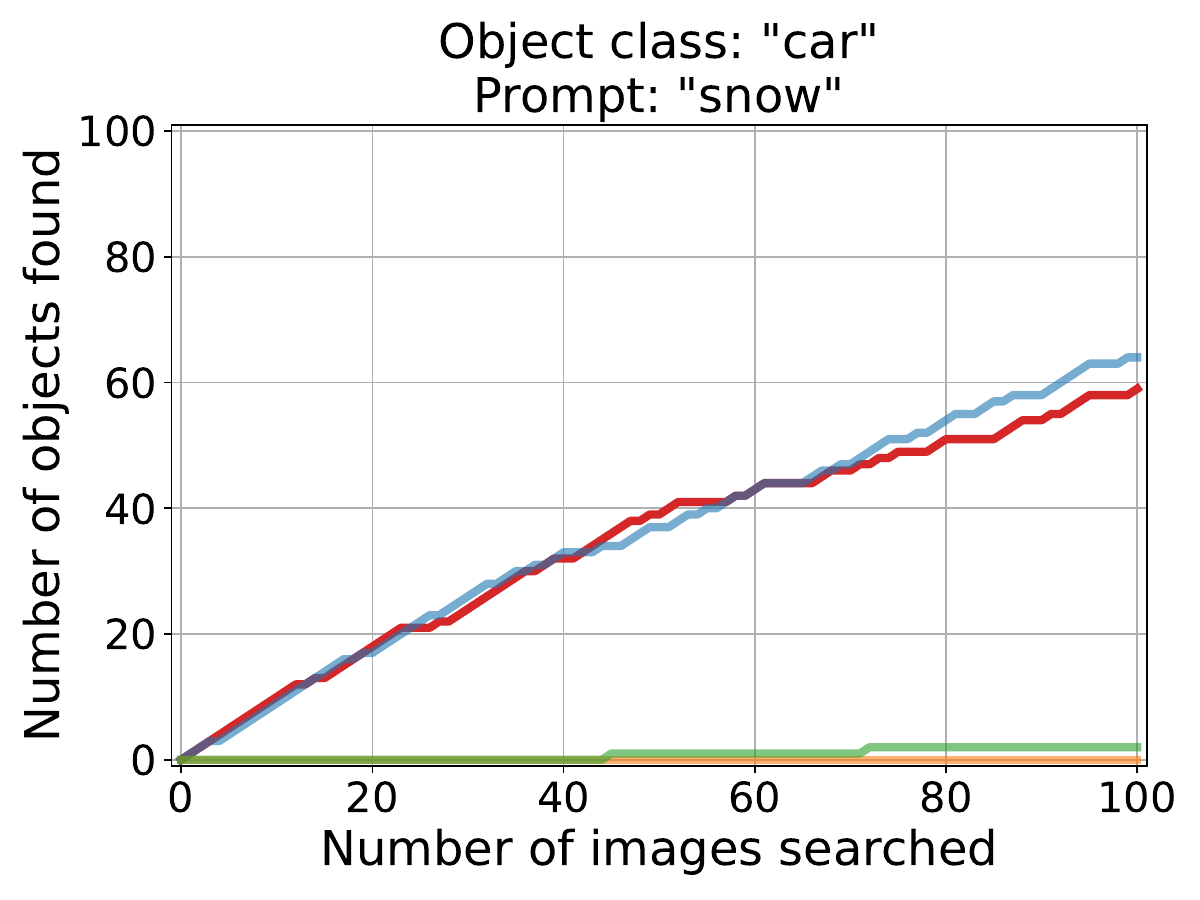}
        \label{fig:image_search_experiment:f}
    \end{subfigure}
    \hfill
    \begin{subfigure}[t]{0.32\textwidth}
        \centering
        \includegraphics[width=\linewidth]{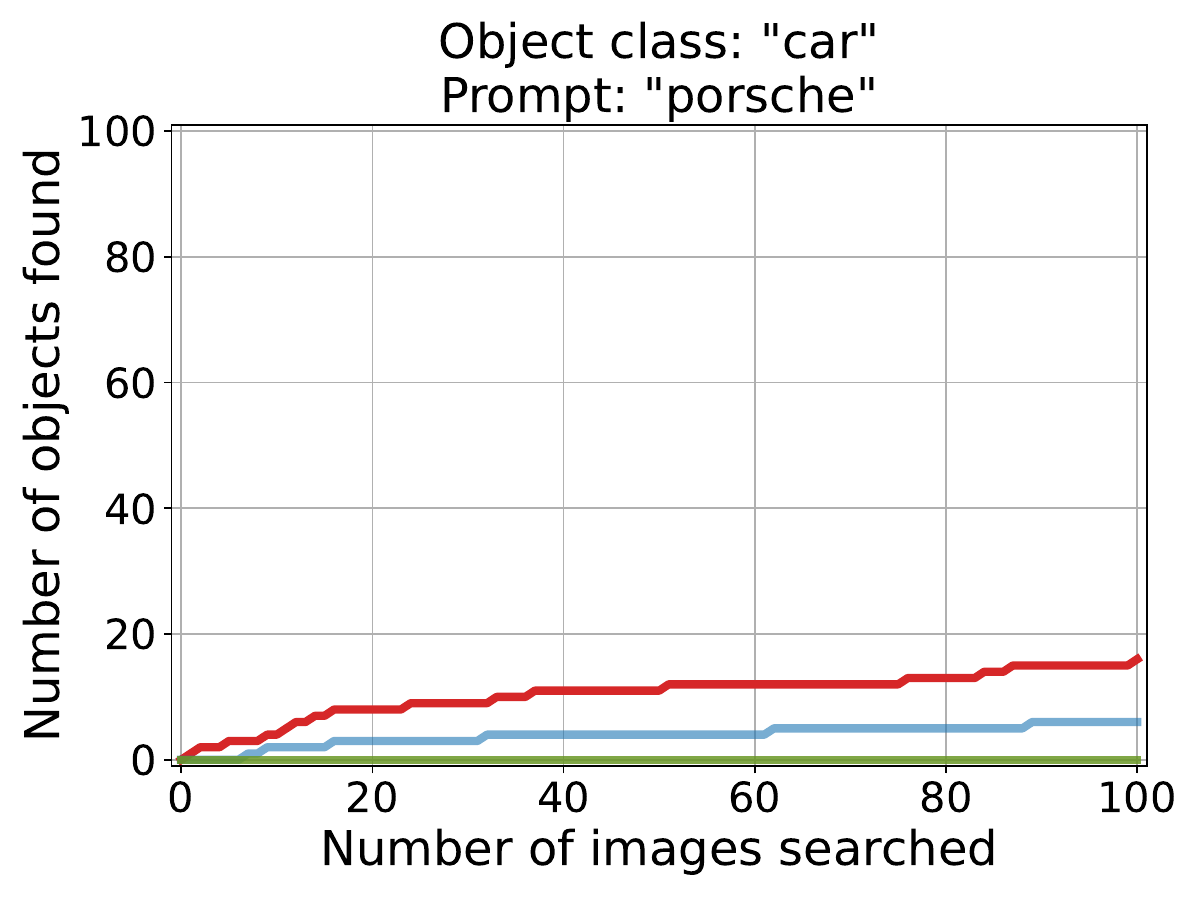}
        \label{fig:image_search_experiment:g}
    \end{subfigure}
    \hfill
    \begin{subfigure}[t]{0.32\textwidth}
        \centering
        \includegraphics[width=\linewidth]{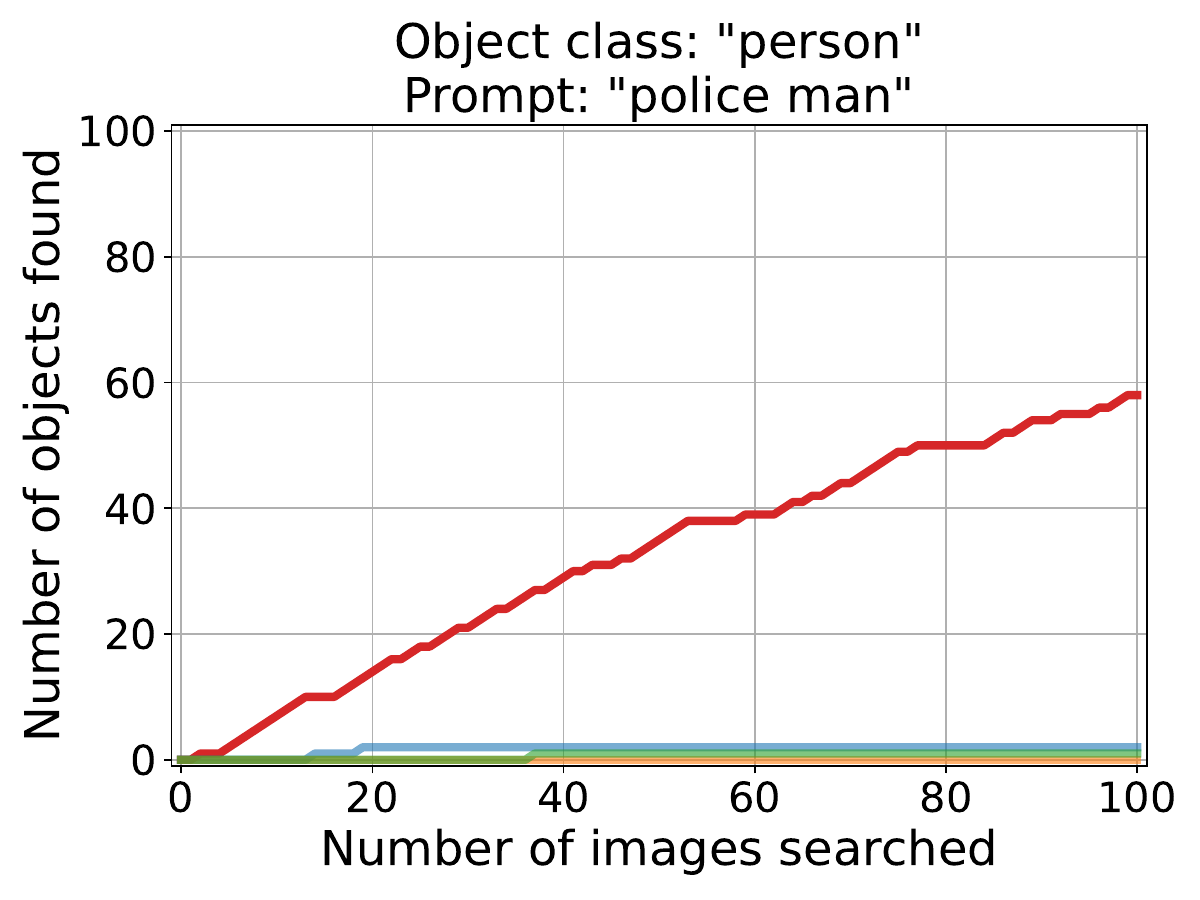}
        \label{fig:image_search_experiment:h}
    \end{subfigure}
    \hfill
    \begin{subfigure}[t]{0.32\textwidth}
        \centering
        \includegraphics[width=\linewidth]{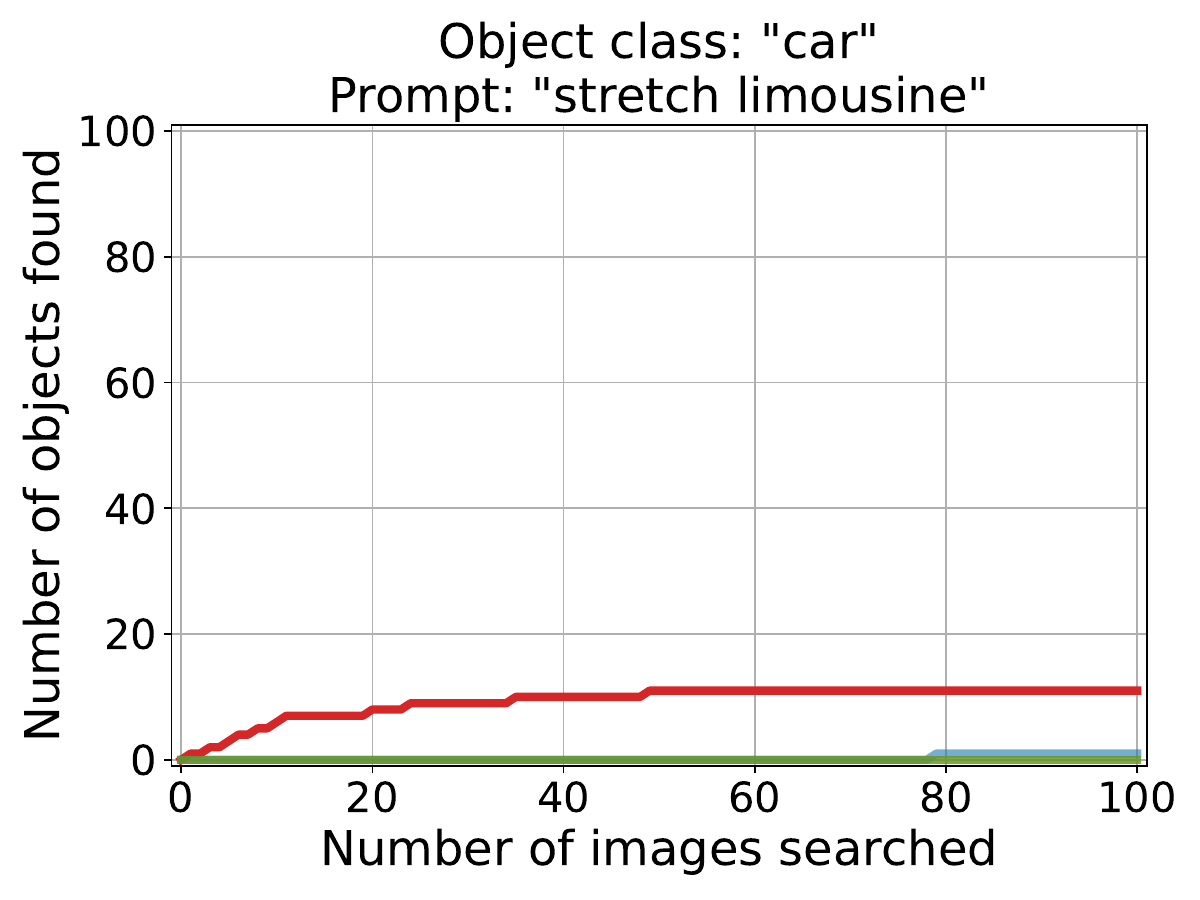}
        \label{fig:image_search_experiment:i}
    \end{subfigure}
    \caption{Cumulative number of true positives with different queries for SOLA (panoptic segmentation + CLIP) and three baseline methods.}%
    \label{fig:image_search_experiment}
\end{figure*}

We evaluated the performance gain of SOLA also on the BDD100k data set~\cite{yu_bdd100k_2020}.
The goal was to find the most diverse objects that could pose a challenge for an automated driving system.
Therefore, we defined some queries $(c_\text{query}, T_\text{query})$ for specific objects with a defined property.
Intentionally, we wanted to compare the performance for probably frequent objects like stop signs and for probably rare objects like stretch limousines.
As a baseline, we have chosen three search strategies for these objects:
\begin{enumerate}
    \item CLIP applied to full image, as in our previous work, and checked if they contain the desired object $c_\text{query}$ with the defined property $T_\text{query}$..
    \item Randomly selected images, which were manually checked if they contain the desired object $c_\text{query}$ with the defined property $T_\text{query}$.
    \item Images preselected by object class $c_\text{query}$ (derived from the panoptic segmentation) and manually checked if they match the defined property $T_\text{query}$.
\end{enumerate}
These three baselines were compared to SOLA.
For the application of CLIP on the full image, we needed to extend some prompts to also include the object class.
The extended prompts are: ``person red top'', ``traffic sign stop'', ``traffic sign parking'', and ``car snow''.
The evaluation of the first two baselines took much longer because the full image had to be searched.
Whereas the third variant was limited to a single object per searched image.

We executed SOLA with the prompt and sorted the data set by the cosine similarity score.
Subsequently, we manually checked the first~100 results with the biggest cosine similarity score for the desired result.
In the evaluation of all three methods, the searched object was assumed to be found only if it could be identified without doubt.
An example of the objects found with SOLA is shown for police men in Fig.~\ref{fig:example_police_man}.

To compare the results, we plotted the cumulative number of true positives for the first~100 images (see Fig.~\ref{fig:image_search_experiment}).
In~8 out of~9 experiments, SOLA has found more objects than the baseline methods, regardless of the number of images searched.
In the case of the prompt (car, snow), our previous method was more successful.
This could be the case because there are often multiple cars in one image that could be covered with snow.
With the CLIP encoder applied to the full image, all these cars are taken into account during evaluation.
In contrast, SOLA only ever looks at a single car of an image.
It also follows that validation with our previous method takes correspondingly more time.

When searching for stop signs, all objects found with SOLA were actually stop signs.
For the random baseline methods, there were only two in each case.
With our previous method, we only found~$36$ stop signs in the first~100 images.

Only one stretch limousine was found with our previous method.
Both random searches found no stretch limousines at all.
But even with SOLA we only found~$11$ stretch limousines.
This may be due to the fact that there are no more images with stretch limousines in the data set, or that SOLA is not able to identify them.
Therefore, the exact false positive rate is unknown.
Moreover, we note that we do not find the~11 stretch limousines directly in the first~11 results, but they extend to the first~49 results.
That means that there are several false positives in the results of SOLA.
Otherwise, all~11 stretch limousines would have been in the first~11 results.

On average, we found~$31.6$ more objects than our previous method,~$54.6$ more than the random search on image level and~$56.1$ more than the random search on object level.

Looking back at our requirements for SOLA:
\begin{itemize}
    \item No additional annotations are necessary to use SOLA.
    \item With SOLA, we are able to search for properties on the object level.
    \item Requests can be formulated using natural language.
    \item By splitting SOLA into a preprocessing and a subsequent image retrieval step, we achieve a reasonable runtime. CLIP only has to perform one inference step during the search, and its runtime is in the millisecond range~\cite{rigoll_focus_2023}. The search in a vector database with millions of entries can also be carried out within milliseconds~\cite{rentong_guo_milvus_2023}.
\end{itemize}

\section{CONCLUSION}
\label{sec:conclusion}
In this work, we addressed the problem of searchability of automotive image data sets at the object level.
Finding objects with specific properties in ever-growing data sets is an important capability for the development and the testing of automated driving systems.
Our goal was to get by without additional annotations and still provide an intuitive search method.
To achieve this, we built an image retrieval tool utilizing state-of-the-art neural networks and extensively evaluated the performance on automotive data sets.
Overall, we have shown that SOLA is a helpful tool, which improves the performance when searching automotive data sets.
The results of SOLA can be understood as candidate proposals for the searched object.
However, due to the preselection, the effort to filter out false positives is minimal.
In addition, it should also be mentioned that false negatives cannot be ruled out.
One additional limitation is, that a targeted search is only possible for object classes which are covered by the panoptic segmentation.
Despite these limitations, SOLA allows a speed increase in the development of automated driving systems by finding objects of interest even in image data sets with tens of thousands of images.
This is especially true in cases, where false positives are tolerable.

To further increase the search performance, our future research will focus on prompt optimization.
This way, we will support a user in crafting precise prompts, and we also hope to reduce false positives.
Furthermore, we will analyze the segmentation networks mentioned in the related work part for their suitability for our problem and find out if we can use them with a reasonable runtime in a similar method as SOLA.

\section*{ACKNOWLEDGMENT}
This work results from the just better DATA (jbDATA) project supported by the German Federal Ministry for Economic Affairs and Climate Action of Germany (BMWK) and the European Union, grant number 19A23003H.

We would like to thank Philipp Schenkel for his suggestions regarding the evaluation and his feedback.

\bibliographystyle{IEEEtran}
\bibliography{IEEEabrv,root}

\end{document}